\documentclass[]{fairmeta}

\title{Interference Matrix: Quantifying Cross-Lingual Interference in Transformer Encoders}

\author[1,2,*]{Belen Alastruey}
\author[1,3,*]{João Maria Janeiro}
\author[2]{Alexandre Allauzen}
\author[1]{Maha Elbayad}
\author[1]{Loïc Barrault}
\author[1]{Marta R. Costa-jussa}

\affiliation[1]{FAIR at Meta}
\affiliation[2]{Université Paris Dauphine - PSL, Paris}
\affiliation[3]{Sorbonne Université, CNRS, ISIR, F-75005 Paris, France}

\contribution[*]{Equal Contribution.}

\abstract{
In this paper, we present a comprehensive study of language interference in encoder-only Transformer models across 83 languages. We construct an interference matrix by training and evaluating small BERT-like models on all possible language pairs, providing a large-scale quantification of cross-lingual interference. Our analysis reveals that interference between languages is asymmetrical and that its patterns do not align with traditional linguistic characteristics, such as language family, nor with proxies like embedding similarity, but instead better relate to script.
Finally, we demonstrate that the interference matrix effectively predicts performance on downstream tasks, serving as a tool to better design multilingual models to obtain optimal performance.

}

\date{\today}
\correspondence{Belen Alastruey at \email{alastruey@meta.com}}

\begin{document}

\maketitle

\section{Introduction}
\label{section:introduction}

With growing interest in models that support a large number of languages, the field of multilinguality is expanding rapidly in many directions,
including large language models (LLMs), encoders, and translation models. 
Therefore, researchers must address the challenges inherent in combining a large number of languages when building massively multilingual models.

Recent progress in Natural Language Processing (NLP) has been driven primarily by a single approach: collecting large multilingual datasets to train transformer models. Models like XLM-R~\citep{conneau2020unsupervised} and mBERT~\citep{devlin-etal-2019-bert}, both encoder-only models, are strong examples of this approach.
These models are often trained with large amounts of multilingual data, but without a deep understanding of the challenges and implications of language sharing and interference.

Understanding how languages interact inside these models is critical for effective multilingual representation learning, and research has already led to some valuable insights. For example, studies on encoder-decoder models for Machine Translation (MT)~\citep{shaham-etal-2023-causes}, and work on various encoder-only architectures~\citep{conneau2020unsupervised}, have often shown that models can effectively share knowledge between similar languages. However, when languages are very different, negative interference hinders the learning process. 
Still, many of these studies have focused on a small set of languages, which makes it unclear if their findings generalize, and which languages cause more or less interference in large multilingual scenarios. 
Hence, a deep understanding of language interaction in multilingual models and what causes transfer or interference is still an open question.

A separate line of research has attempted to approximate interference between languages, through different strategies such as language families~\citep{tan-etal-2019-multilingual}, embedding similarity~\citep{raganato-tiedemann-2018-analysis} or gradients~\citep{wang-etal-2023-gradsim}. However, these methods are estimations and it is unclear how well they represent actual model behavior.

In this work, we aim to conduct a large scale study of how languages interfere with each other in encoder-only Transformer models, and how this information can be used to improve the performance of multilingual models. To do so, we build an interference matrix by training and evaluating small BERT-like models on all possible language pairs across 83 languages. 

Subsequently, we analyze the results obtained from the interference matrix to identify potential causes of higher interference and determine which languages are more likely to degrade or \emph{harm} a model. 
Finally, we train standard-sized BERT models and evaluate them on the Massive classification tasks~\citep{fitzgerald2022massive} through the MTEB benchmark. Our findings demonstrate that the interference scores from the matrix can effectively predict performance drops in downstream tasks and help identify languages that degrade the overall performance in multilingual models.

\noindent Our main contributions include:
\begin{itemize}
    \item We introduce an \emph{Interference Matrix}, derived from a large-scale analysis of 83 languages, that quantifies cross-lingual interference in encoder-only transformers.
    
    \item We show that not all languages converge equally well, even when trained under identical conditions.
    
    \item We reveal that cross-lingual interference is \emph{asymmetric}; i.e., the impact of language A on B is not the same as the impact of language B on A.
    
    \item We find that interference is independent of common proxies such as embedding similarity and language family, but is correlated with the languages' script.
    
    \item We observe that low-resource languages are more susceptible to performance drops and are more likely to negatively impact the performance of other languages.
    
    \item We show that our matrix can effectively predict performance in downstream tasks, serving as a tool to guide the design of stronger multilingual models.
\end{itemize}

\section{Related Work}

The field of multilingual NLP has made notable progress with the introduction of large, fully shared multilingual encoders such as mBERT~\citep{devlin2019bert} and XLM-RoBERTa~\citep{conneau2020unsupervised}. 
These models have demonstrated an effective approach to learning multilingual representations by pretraining a single shared encoder on extensive, mixed-language datasets using self-supervised learning objectives. 
Despite their success, these multilingual models face a significant challenge known as the \textit{``curse of multilinguality''}, where the model's performance deteriorates as the number of languages increases due to growing interference among languages. This phenomenon has motivated researchers to understand how languages interact with each other and how to prevent interference.

\textbf{Language specific layers.} One approach to mitigate these issues is the use of language-specific layers. In the area of machine translation (MT), researchers have thoroughly investigated this approach~\citep{pires-etal-2023-learning, pfeiffer2023mmt5modularmultilingualpretraining, qu2025languagestransferredencoderrepresentation, purason-tattar-2022-multilingual}, with some efforts also focusing on language-specific components within the encoder~\citep{qu2025languagestransferredencoderrepresentation}. These studies highlight that introducing parameters specific to individual languages or language groups can lead to enhanced performance in machine translation tasks, by reducing interference between languages.

\textbf{Language interference Approximation.} Historically, language interaction has been approached from different perspectives. \citet{pichel-campos-etal-2018-measuring} have computed language distances using perplexity.  Some studies utilize language family taxonomies~\cite{tan-etal-2019-multilingual, shaffer-2021-language-clustering}, while others represent languages as information-dense vectors that reflect typological or conceptual characteristics~\cite{littell-etal-2017-uriel, lin-etal-2019-choosing, oncevay-etal-2020-bridging}. Another research direction computes language similarity using embeddings derived from multilingual pretrained language models~\cite{raganato-tiedemann-2018-analysis, lange-etal-2021-share, tan-etal-2019-multilingual, shaffer-2021-language-clustering}. 
In a different approach, \citet{wang-etal-2023-gradsim} proposes estimating language similarity for grouping by using the cosine similarity of gradient updates during model optimization.
In a study closely related to ours, \citet{protasov-etal-2024-super} consolidate the concept of language donors and recipients by studying how continued pretraining on different high-resource languages affects the performance in representing low-resource languages.

\textbf{Our contribution in context.} Expanding on previous research on language interference we measure the interference of 83 languages on small bilingual encoder-only models. Unlike previous work that approximates language similarity using various strategies, our approach involves constructing an interference matrix that quantifies cross-lingual interference directly from the model loss. Our findings offer new insights into the ``curse of multilinguality'' and reveal that previously used approximation methods are not reliable proxies for actual model interference.

\section{Interference Matrix}
\label{section:interference_matrix}

Our goal is to perform an in-depth analysis of how languages interact with each other, and how this information can be used to build better multilingual models.
To understand the interaction between languages, we start by selecting a large set of languages and 
we train two types of small Transformer encoders: baseline monolingual models for each language, and a complete set of bilingual models covering every possible language pair.
We then evaluate the performance of each model on its respective language(s), and build the Interference Matrix, where it is possible to see the impact each language has on every other language, when combined within a bilingual model.

\subsection{Experimental Setup}
\label{section:experimental_setup}
In order to build our matrix, we train a BERT-like model for each language and each language pair. To make this large amount of experiments feasible, and to amplify the consequences of mixing each pair of languages, we choose to train shallow 2-layer transformer encoders. 

Our model implementation consists of a modification of the XLM-R implementation from HuggingFace, to include the memory efficient attention from xformers~\citep{xFormers2022}.

We train our models using the XLM-R tokenizer and a subset of the data from NLLB~\citep{nllbteam2022languageleftbehindscaling}, in particular we select 1M samples for each language. 
We perform this analysis on 83 languages, which correspond to the intersection of languages covered by the XLM-R tokenizer and those with at least 1M samples available in the NLLB dataset. The list of the selected 83 languages is available in Appendix \ref{appx:languages}.

We evaluate the models on FLORES200 dataset devtest set~\citep{nllbteam2022languageleftbehindscaling} for each language separately.

We train the models for 10,000 steps using a learning rate of $1\times10^{-5}$, with warmup for 2,500 steps and a cosine decay for the remaining steps.
We use the AdamW optimizer~\citep{loshchilov2018decoupled}.
The training objective is Masked Language Modeling (MLM) with single sentence, as proposed in BERT~\citep{devlin-etal-2019-bert}.
We set a maximum sequence length of 256 tokens, and use a masking ratio of 15\%, as done in BERT.

\begin{figure*}
    \centering
    \includegraphics[width=\linewidth]{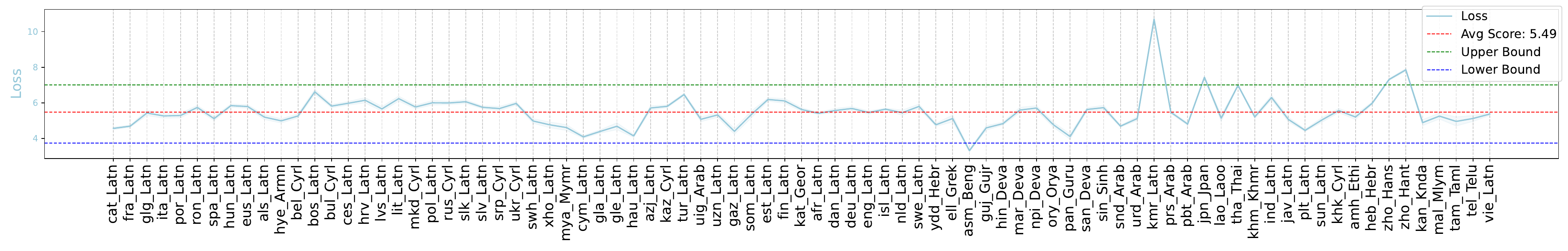}
    \caption{Average loss ("Loss") obtained across models that cover a language (average of matrix $L$ row-wise). The average score, and lower and upper bounds are obtained through statistical analysis of the loss curve and used to understand which languages are outliers.}
    \label{fig:loss}
\end{figure*}

\subsection{Model Convergence by Language}
\label{sec:loss}
Before building the interference matrix we first analyze the convergence of the trained models. Each monolingual and bilingual model is tested on the corresponding FLORES200 set for the one or two languages it covers. 
Using these scores, we create a loss matrix $L$ (can be found in Appendix~\ref{appx:matrices}, Figure~\ref{fig:loss_matrix}) where each row corresponds to a language in FLORES200 devtest set and each column indicates the secondary language used in the model.
To aggregate the results, we average the scores on each devtest set (average of $L$ by rows) and present the results in Figure \ref{fig:loss}.

Interestingly, even though all languages are trained with the same amount of resources and the models have the same sizes, we observe significant disparities in how different languages behave. This suggests that some languages are easier to model than others, meaning the amount of data and model capacity required to optimally train a model varies across languages. This behavior could be a result of different causes, such as data quality, tokenizer coverage of the language, or differences on the numbers of tokens in each language (only the number of sentences is fixed).

Furthermore, we perform a statistical analysis of the scores to identify outliers. We employ the Interquartile Range (IQR) method, calculating the first and third quartiles ($Q1$, $Q3$) and the $IQR$ ($Q3-Q1$). 
Any data point falling outside the bounds of $[Q1 - 1.5 \times IQR, Q3 + 1.5 \times IQR]$ is considered an outlier.
We plot these bounds and the average loss in Figure \ref{fig:loss}, and determine that Kurdish, Japanese and both simplified and traditional Chinese are outliers that converged poorly, indicating these are hard to fit languages in this experimental setup.

\begin{figure*}
    \centering
    \includegraphics[width=0.7\linewidth]{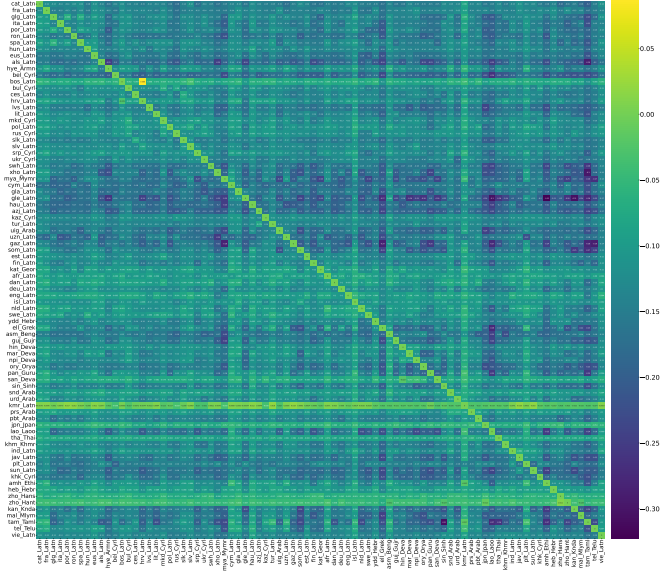}
    \caption{Interference Matrix.  By rows, we see the impact in a language of adding additional languages in the training mix. By columns, we see the impact of adding a specific language on the performance of all other languages.}
    \label{fig:interference_matrix}
\end{figure*}

\subsection{The Interference Matrix}
Using the loss matrix $L$, we build the interference matrix $I$.
We denote $L_{\text{lang}_A}$ as the loss of the model trained solely on language~A, and $L_{\text{lang}_A,\text{lang}_B}$ as the loss on language A of a model trained on both languages~A and B. The interference matrix $I$ is then derived from the loss matrix $L$ as follows:

\begin{equation}
    I_{\text{lang}_A, \text{lang}_B} = \frac{L_{\text{lang}_A} - L_{\text{lang}_A, \text{lang}_B}}{L_{\text{lang}_A}},
\end{equation}
where $I_{\text{lang}_A, \text{lang}_B}$ represents the interference that language B causes on language A. 
Since adding a language typically increases loss ($L_{\text{lang}_A, lang_B} > L_{\text{lang}_A}$), the numerator is generally negative.
Therefore, a more negative (i.e., lower) score in the matrix indicates a larger performance drop and thus higher interference.

When analyzing the full interference matrix $I$, a specific row for language~A shows how every other language affects its performance. A language with high scores in its row (i.e., small loss differences) is considered \emph{robust}, while a language with low scores in its row (i.e., high loss differences) is considered \emph{weak}. 
The row average quantifies a language's overall robustness.
Reading by columns shows the impact of adding a specific language on the performance of all other languages. 
A language with low scores in its column (i.e., high loss differences) has a strong \emph{damaging} effect, while a language with high scores (i.e. small loss differences) is considered \emph{friendly} to others. The column average quantifies a language's friendliness.

In Figure \ref{fig:interference_matrix} we present the full matrix sorted by linguistic family. 
We observe that all results in the matrix but one are negative scores, meaning that positive transfer of information is rare at this scale. 
The only language that slightly benefited from being mixed is Bosnian (bos\_Latn) when mixed with Croatian (hrv\_Latn), likely due to their high similarity.
Furthermore, we observe stripped patterns, both row- and column-wise, meaning that some specific languages show higher robustness or friendliness, and in particular, we observe that the stripes seem to align with linguistic families.

In the following sections, we analyze the matrix results across different criteria. To avoid biased results, we exclude the outliers detected in Section \ref{sec:loss} from this analysis.


\begin{figure*}
    \centering
    \includegraphics[width=1\linewidth]{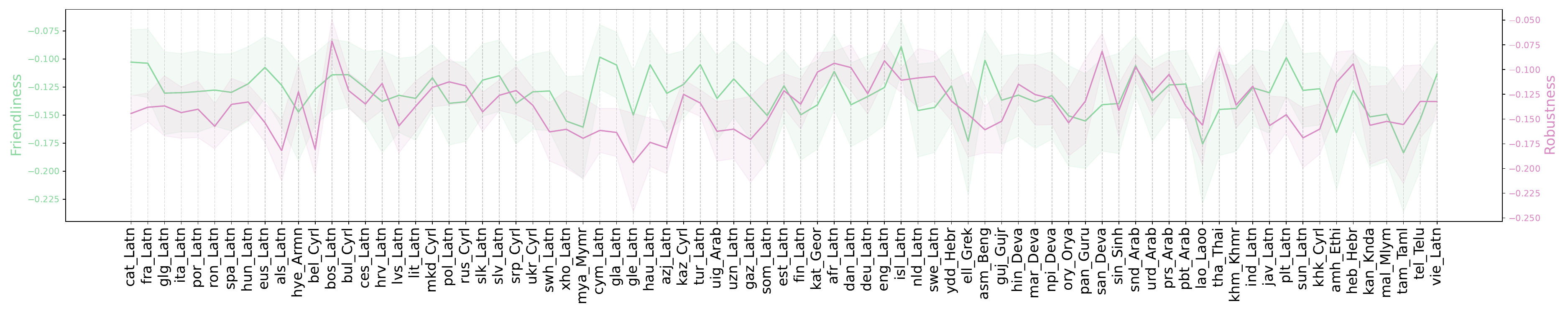}
    \caption{Friendliness and robustness of all studied languages. We observe dissimilar patterns between robustness and friendliness.}
    \label{fig:harm_vs_weakness}
\end{figure*}

\begin{figure}[ht]
    \centering
    \begin{minipage}[t]{0.48\textwidth}
        \vspace{0pt} 
        \centering
        \includegraphics[width=\linewidth]{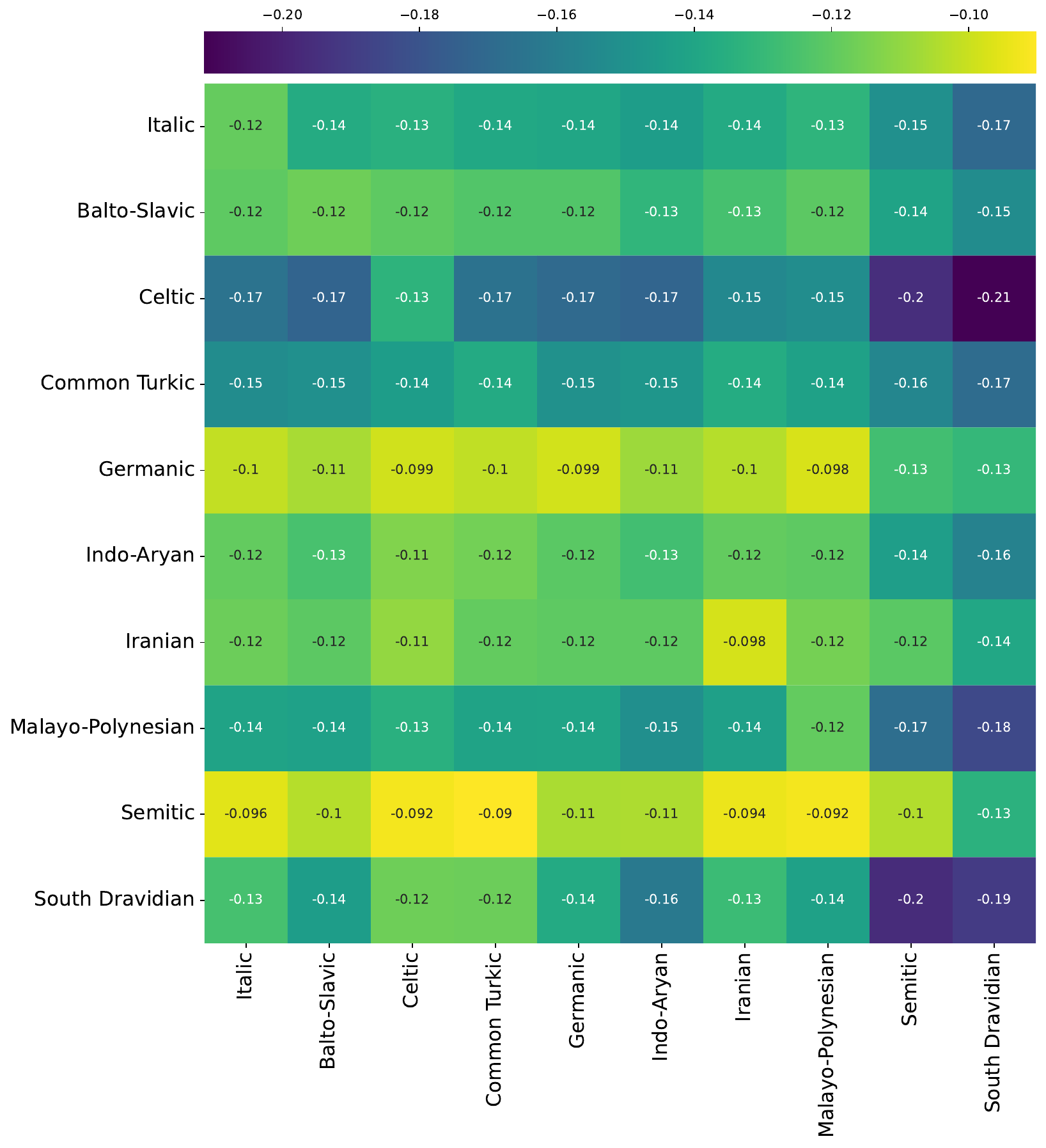}
        \caption{Interference Matrix averaged by linguistic family. In each entry we see the expected loss degradation if mixing languages from different or the same family.}
        \label{fig:interference_by_language}
    \end{minipage}
    \hfill
    \begin{minipage}[t]{0.425\textwidth}
        \vspace{0pt} 
        \centering
        \includegraphics[width=\linewidth]{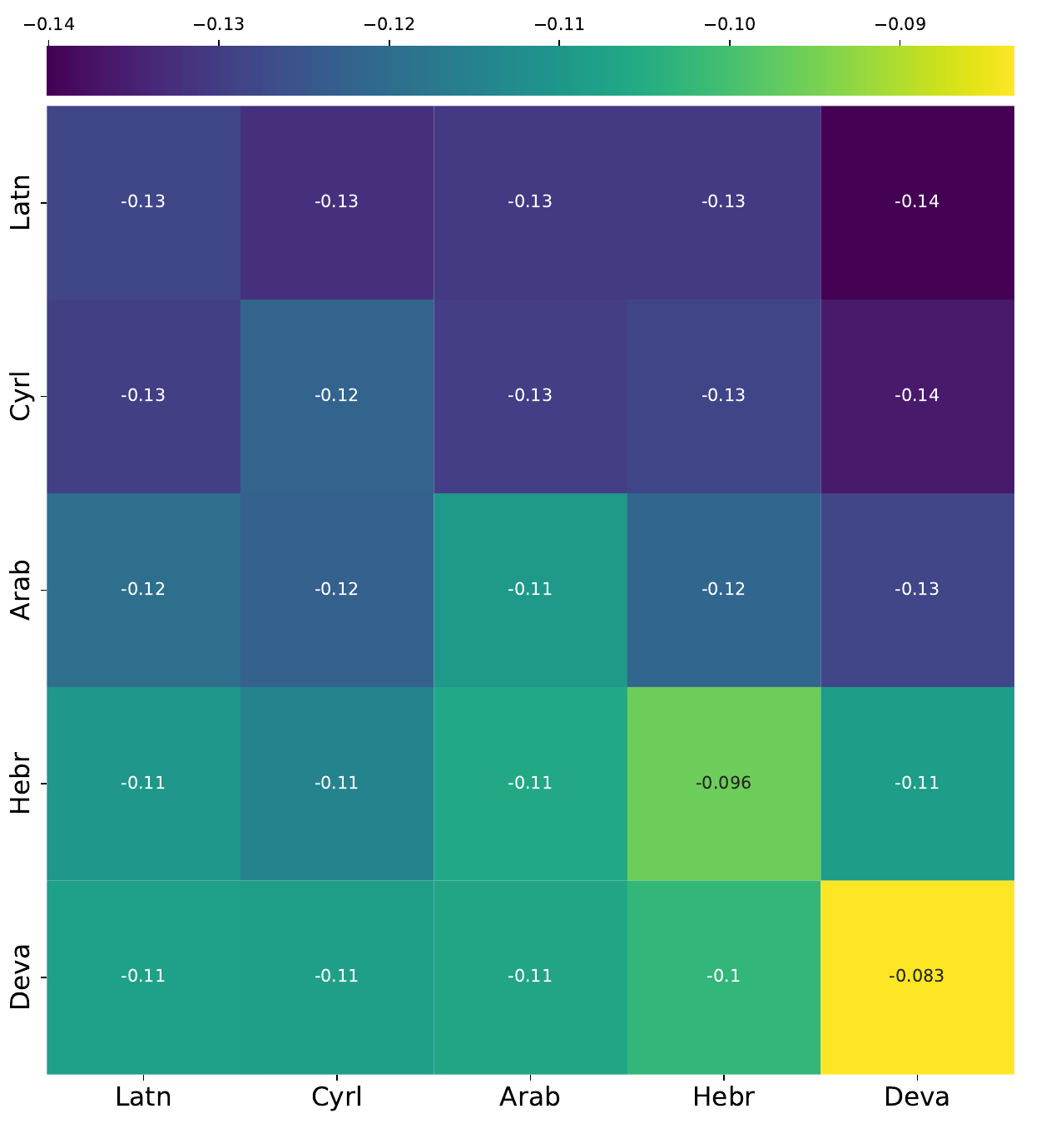}
        \vspace{17pt} 
        \caption{Interference Matrix averaged by script. In each entry we see the expected loss degradation if mixing languages from different or the same script.}
        \label{fig:interference_by_script}
    \end{minipage}
\end{figure}

\subsection{Robustness vs. Friendliness}
\label{sec:robvsfri}
A first intuitive analysis of the matrix consists of understanding the relationship between the robustness and friendliness of a language. 
We compute the average of the scores in the matrix both row-wise (robustness) and column-wise (friendliness). As shown in Figure \ref{fig:harm_vs_weakness}, these two variables do not seem to follow a similar trend.

In particular, we also observe that the matrix is not symmetrical. For example, Welsh (cym\_Latn) appears friendly to other languages but is not itself robust. 
This highlights that interference is an \emph{asymmetric, directional concept}, rather than a single symmetric value.
The interference that language A produces on language B can be different from the interference that language B causes on language A.

\subsection{Language Families and Scripts}
\label{sec:family}

In this section, we aim to understand whether languages belonging to the same linguistic family or written in the same script exhibit distinct interference patterns. To investigate this, we consider the linguistic families described in NLLB~\citep{nllbteam2022languageleftbehindscaling}, focusing on families with at least three languages represented in our interference matrix. 
We then average all scores in the corresponding regions of the matrix defined by these families (excluding the zeros on the diagonal) to obtain the matrix shown in Figure \ref{fig:interference_by_language}.

In this matrix, each entry represents the expected interference experienced by a language of a given family (row) when mixed with a language from another family (column). A common assumption in NLP is that languages from the same linguistic family provide mutual benefits, which would imply higher values (less interference) along the diagonal. 
However, our results do not show this pattern. Instead, we observe distinct row-wise patterns indicating that some families are generally more robust.
For example, Semitic and Germanic languages appear highly robust against interference,
whereas Celtic and Common Turkic languages are more susceptible to interference. 
Additionally, Semitic and South Dravidian families tend to cause more harm to other languages.

We conduct a similar analysis based on writing scripts with results shown in Figure \ref{fig:interference_by_script}. 
Here, a clearer diagonal pattern emerges, suggesting that languages sharing a script interfere less with each other. 
Moreover, languages using Cyrillic and Latin scripts appear less robust compared to those written in Hebrew or Devanagari scripts.

\begin{figure}[]
  \centering
  \begin{minipage}[b]{0.25\linewidth}
    \centering
    \includegraphics[width=\linewidth]{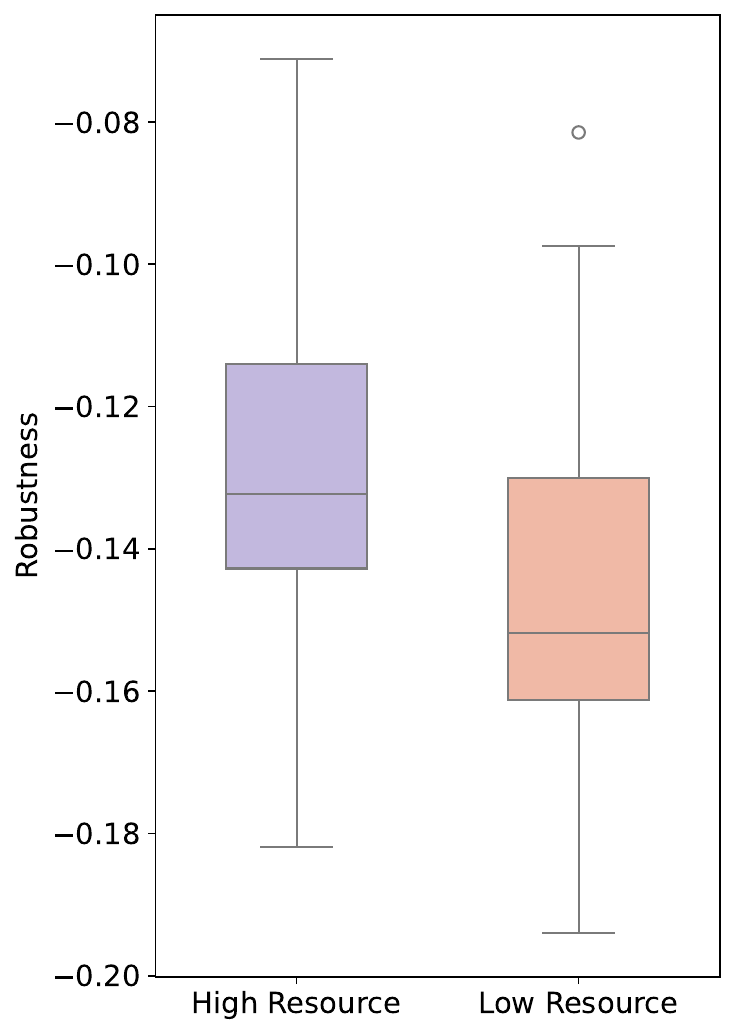}
  \end{minipage}%
  \hspace{0pt}%
  \begin{minipage}[b]{0.25\linewidth}
    \centering
    \includegraphics[width=\linewidth]{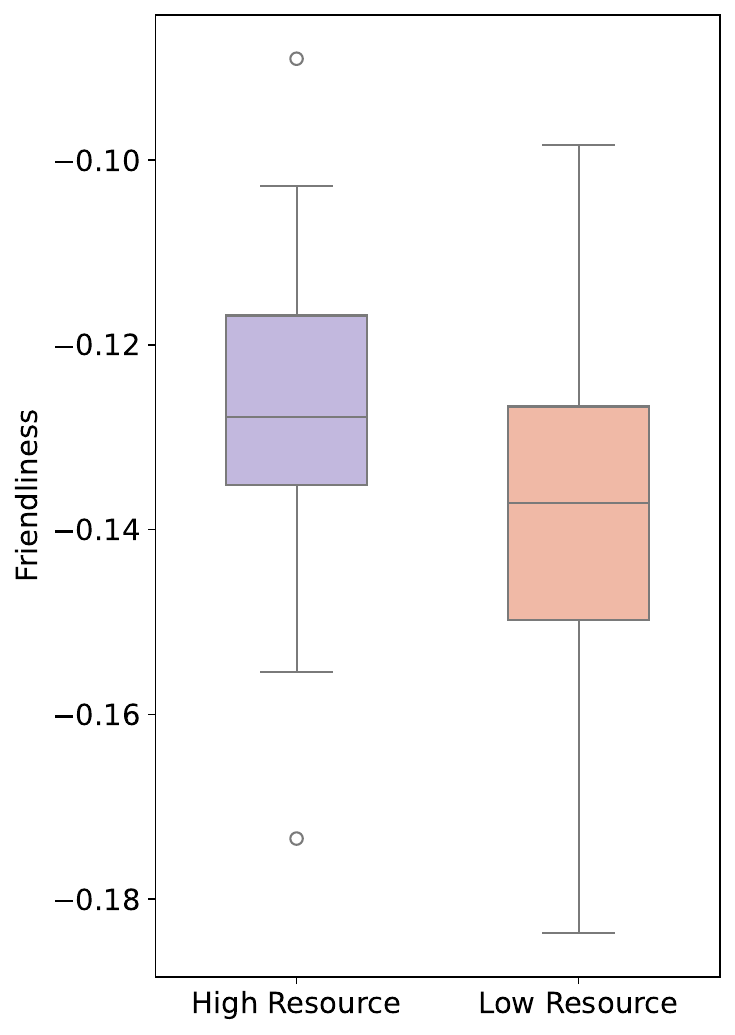}
  \end{minipage}
  \caption{Average Robustness and Friendliness grouped by language resources.}
  \label{fig:high_low_resource}
\end{figure}

\subsection{High- and Low-Resource Languages.}
\label{sec:resources}
Furthermore, we explore whether a language's resource level correlates with its robustness and friendliness.
Although we allocate the same amount of training data to each language in our experimental setup, the underling data quality can still vary between high- and low-resource languages. 
To analyze this, we use the resource classification from the NLLB paper. 
The results presented in Figure~\ref{fig:high_low_resource} show that low-resource languages tend to exhibit both lower friendliness and lower robustness than high-resource languages. 
This behavior can likely be attributed to data quality; since models trained on lower quality data are harder to fit, their performance is more susceptible to degradation when a second language is introduced.
This difficulty in fitting data of the primary language can, in turn, negatively impact the model's ability to learn the secondary language, i.e. it has a larger damaging effect on the language it is combined with.

\begin{figure*}
    \centering
    \includegraphics[width=\linewidth]{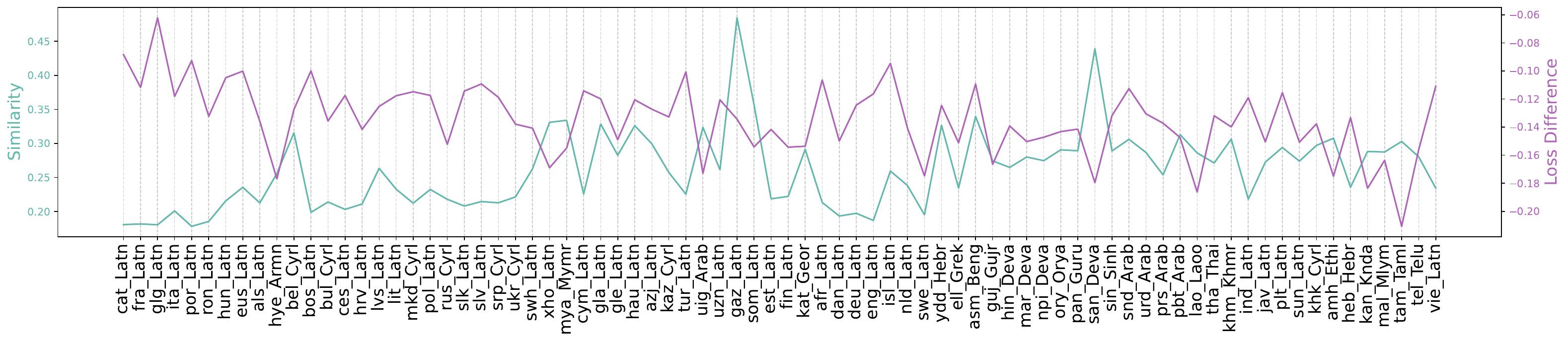}
    \caption{Comparison between the row corresponding to Spanish of the interference matrix and MEXMA similarity. We conclude that the trends don't have a relationship.}
    \label{fig:similarity}
\end{figure*}

\section{Correlation with Embedding Similarity}

In this section, we aim to explore whether alternative approximation methods that do not require training a large number of models can be used to estimate the interference between languages. Specifically, we investigate embedding similarity using SONAR~\citep{duquenne2023sonarsentencelevelmultimodallanguageagnostic} and MEXMA~\citep{janeiro-etal-2025-mexma} embeddings.

For this analysis, we use the FLORES200 devtest set. We embed all translation pairs using both SONAR and MEXMA encoders and then calculate the similarity between any two languages by averaging the cosine similarities of the embedded translation pairs. 
From these scores, we build language similarity matrices for each encoder.
A key structural difference arises immediately; these similarity matrices are symmetric whereas our interference matrix $I$ is asymmetric.

To test for similar patterns between these matrices, we compare row by row the interference matrix and each embedding similarity matrix and observe if languages that are closer in the embedding space (higher similarity) experience less interference.
Figure \ref{fig:similarity}, which shows the results for Spanish using MEXMA, reveals that the two metrics follow different trends. This lack of correlation holds for other languages and for the SONAR encoder as well (see Appendix \ref{appx:embedding_similarity}), suggesting that embedding similarity is not a reliable proxy for predicting directional interference between languages.

\section{Downstream Evaluation}
In this section we focus on understanding how the differences in interference we observe in our matrix affect standard-sized models in downstream tasks.
\begin{table*}[!t]
\centering
\small
\begin{tabular}{lc|cc|c|cc|c|c}
\toprule
\multicolumn{2}{c}{} & \multicolumn{3}{c}{\textbf{Low Interference}} & \multicolumn{3}{c}{\textbf{High Interference}} & \\ 
\cmidrule(lr){3-5}
\cmidrule(lr){6-8}
 & \textbf{ell} & \textbf{ell-kmr} & \textbf{ell-plt} & \textbf{Avg} & \textbf{ell-tam} & \textbf{ell-mal} & \textbf{Avg} & $\Delta$\\ 

\midrule
\textbf{Massive Intent}   & 45.78 & 43.32 & 42.07 & 42.69 & 39.87 & 40.53 & 40.20 & 2.48 \\
\textbf{Massive Scenario} & 49.17 & 44.69 & 43.53 & 44.11 & 39.73 & 40.77 & 40.25 & 3.86 \\
\toprule
 & \textbf{mya} & \textbf{mya-plt} & \textbf{mya-fra} & \textbf{Avg} & \textbf{mya-tel} &\textbf{mya-tam}  & \textbf{Avg} & $\Delta$\\
 \midrule
 \textbf{Massive Intent} & 52.89 & 51.07 &  50.88 & 50.97 &49.88 & 50.54 & 50.21 & 0.76 \\
\textbf{Massive Scenario} & 56.72 & 54.81 & 54.75 & 54.78 &54.10 & 53.84 & 53.97 & 0.81 \\

\toprule
 & \textbf{nld} & \textbf{nld-hau} & \textbf{nld-fra}& \textbf{Avg} &\textbf{nld-amh} &\textbf{nld-xho} & \textbf{Avg} & $\Delta$\\
 \midrule
\textbf{Massive Scenario} & 48.02 & 40.64 & 40.77 & 40.70 & 39.92 & 40.08 & 40.00 & 0.70 \\
 \textbf{Massive Intent}  & 44.56 & 43.32 & 43.68 & 43.50 & 40.98 & 42.11 & 41.55 & 1.95 \\

\bottomrule
\end{tabular}
\caption{Interference in bilingual trainings in downstream tasks. For each studied language (ell, mya, nld), we evaluate the performance of a model trained on that language paired with either a high-interference or low-interference language. We conduct experiments using two high-interference and two low-interference languages for each studied language. In the columns labeled "Avg," we report the average performance for the high- and low-interference language pairs, and in the "$\Delta$" column, we show the performance drop between high and low interference conditions.}
\label{tab:dowsntream_eval_bilingual}
\end{table*}

\subsection{Experimental Setup}
For these experiments, we extend our training set to 20M sentences per language and train 24-layer models, similar to the original XLM-R model \cite{xlm-r}. The models are now trained for 100,000 steps. We also update the training strategy to use FSDP2~\cite{PyTorch_FSDP2_Tutorial} instead of DDP, with fp32/bf16 mixed-precision instead of fp32/fp16. Other training details remain as described in Section~\ref{section:experimental_setup}.

\subsection{Evaluation}
\label{sec:eval}
For evaluation, we consider two MTEB classification tasks: \texttt{MassiveIntentClassification} and \texttt{MassiveScenarioClassification}~\citep{fitzgerald2022massive}. These tasks involve standard classification of sentence embeddings, and the reported metric is accuracy. 

\subsection{Results in Bilingual Models}
\label{sec:bilingual_exp}
For these experiments, we select three target languages from different scripts and families: Greek (ell\_Grek), Burmese (mya\_Mymr), and Dutch (nld\_Latn). 
For each target language, we identify two languages from our matrix that exhibit high interference and two that exhibit low interference.
We then train monolingual models for each target language and bilingual models for each of the corresponding language pairs. 

The results are presented in Table \ref{tab:dowsntream_eval_bilingual}. The $\Delta$ column shows the performance gap between high- and low-interference conditions when evaluating on the target language.
We observe that for all target languages, and as expected, adding a second language to the training data leads to a degradation in performance compared to the monolingual model.
However, when the added language is one identified as high-interference by our matrix, the performance drop is more pronounced than when adding a low-interference language, showing that the matrix can effectively predict downstream performance.

\subsection{Results in Multilingual Models}
Furthermore, we extend these experiments to multilingual scenarios. 

\textbf{Impact of high- and low- interference groups on a target language.} We expand the bilingual experiments by selecting four high- and four low-interference languages for each target language.
We then train two multilingual models for each target language: one with the low-interference group and one with the high-interference group. 

The results in Table \ref{tab:dowsntream_eval_multilingual}, show again that all models suffer performance drops compared to the monolingual baselines. In this set of experiments, as expected, the drop is higher than in the bilingual setting, given that we add more interfering data. Again, models trained with groups of high-interference languages experience greater performance degradation compared to those trained with low-interference groups, confirming our matrix's predictive power.

\begin{table*}[!t]
\centering
\small
\begin{tabular}{lc|c|c|c}
\toprule
\multicolumn{2}{c}{} & \multicolumn{1}{c}{\textbf{Low Interference}} & \multicolumn{1}{c}{\textbf{High Interference}} & \\ 
\cmidrule(lr){3-4}
 & \textbf{ell} & \textbf{ell-plt-isl-tur-fra} &  \textbf{ell-lao-mal-tam-som} & $\Delta$\\ 
\midrule
\textbf{Massive Intent} & 45.78 & 36.44 &  36.17 & 0.27 \\
\textbf{Massive Scenario} & 49.17 & 36.14 & 35.82& 0.32  \\
\toprule
 & \textbf{mya} & \textbf{mya-plt-kmr-fra-slv} &  \textbf{mya-tel-tam-swe-kat} & $\Delta$\\
 \midrule
 \textbf{Massive Intent} & 52.89 & 46.95 & 46.47 & 0.48 \\
\textbf{Massive Scenario} & 56.72 & 50.01 & 49.58 & 0.43  \\
\toprule
 & \textbf{nld} & \textbf{nld-isl-fra-hau-kon} &  \textbf{nld-lao-tam-amh-xho} & $\Delta$\\
 \midrule
 \textbf{Massive Intent}  & 44.56 & 37.07 & 36.64 & 0.43 \\
\textbf{Massive Scenario} & 48.02 & 37.69 & 37.21 & 0.49 \\
\bottomrule
\end{tabular}
\caption{Interference in multilingual trainings in downstream tasks. For each studied language (ell\_Grek, mya\_Mymr, nld\_Latn), we evaluate the performance of a model trained on that language paired with a group of either a high-interference or low-interference languages. In the "$\Delta$" column, we show the performance drop between high and low interference conditions.}
\label{tab:dowsntream_eval_multilingual}
\end{table*}

\textbf{Impact of adding an unfriendly language to a multilingual setting.} In these experiments, we aim to understand the impact on downstream performance we can expect if we include an unfriendly language in the data mix.
For this purpose, we select four languages: Spanish (spa\_Latn), Dutch (nld\_Latn), Albanian (als\_Latn) and Italian (ita\_Latn). 
These languages have a low level of interference with each other. Then, we choose four additional languages, two that have a friendly behavior with the group, Hausa (hau\_Latn) and English (eng\_Latn), and two that do not, Greek (ell\_Grek) and Lao (lao\_Laoo). 

In Table \ref{tab:unfriendly_language_analysis} we can see that overall languages suffer a performance drop when adding an unfriendly language in the mix. These findings indicate that performance of multilingual models can be boosted by removing specific harmful languages, and languages can be carefully mixed to maximize performance on desired languages.

\begin{table}[!t]
\centering
\small
\begin{tabular}{clcc|c}
\toprule
 & & \textbf{+ hau/eng} & \textbf{+ ell/lao} & $\Delta$ \\ 
\midrule
\multirow{4}{*}{\shortstack{\textbf{Massive} \\ \textbf{Intent}}}& \textbf{spa}   & 37.63 & 36.80 & 0.84 \\
& \textbf{nld} & 37.22 & 37.02 & 0.20 \\
& \textbf{als} & 42.91 & 42.91 & 0.00 \\
& \textbf{ita} & 41.43 & 40.58 & 0.84 \\
\midrule
\multirow{4}{*}{\shortstack{\textbf{Massive} \\ \textbf{Scenario}}}& \textbf{spa} 
               & 39.37 & 37.92 & 1.45 \\
& \textbf{nld} & 39.73 & 39.08 & 0.65 \\
& \textbf{als} & 43.60 & 43.42 & 0.17 \\
& \textbf{ita} & 41.23 & 40.43 & 0.80 \\
\bottomrule
\end{tabular}
\caption{Interference in multilingual trainings in downstream tasks. We evaluate the performance of a model trained on four target languages together with a low or high interfering language. We present the averaged results for Hausa and English, and for Greek and Lao. In the ``$\Delta$'' column, we show the performance drop between high and low interference conditions.}
\label{tab:unfriendly_language_analysis}
\end{table}

\section{Conclusion}
\label{sec:conclusions}
In this work we presented a large-scale study of language interference in encoder-only Transformer models, constructing an \emph{Interference Matrix} to quantify these effects across 83 languages.
Our analysis reveals that interference is asymmetric and that its patterns contradict common assumptions, aligning more strongly with writing script rather than proxies like language family or embedding similarity. 
We also establish that low-resource languages are not only more vulnerable to negative interference but are also more likely to degrade the performance of other languages in a multilingual setting.
We demonstrate our interference matrix is more than an analytical tool; it can effectively predict performance degradation on downstream tasks. By enabling the identification of harmful languages, our work provides a practical methodology to aid the design of better performing multilingual encoders.

\clearpage
\newpage
\bibliographystyle{assets/plainnat}
\bibliography{paper, anthology}

\clearpage
\newpage
\beginappendix

\section{Full Matrices}
\label{appx:matrices}

\begin{figure*}[!h]
    \centering
    \includegraphics[width=0.85\linewidth]{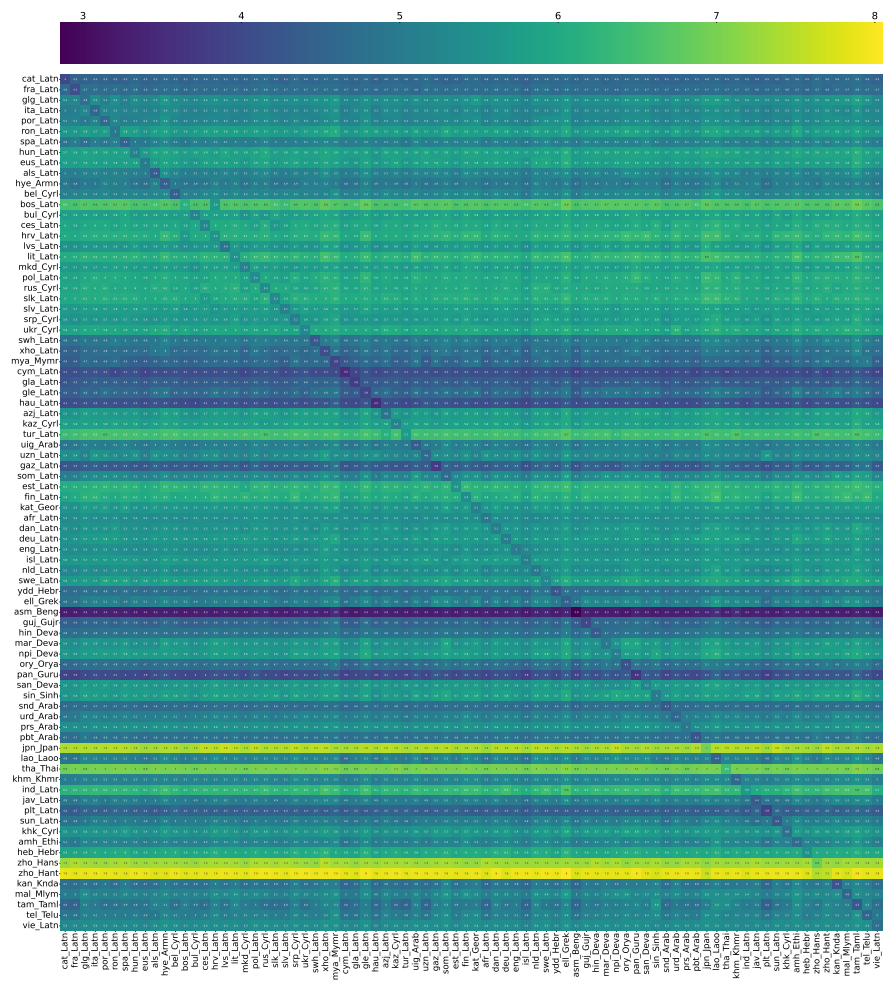}
    \caption{Loss Scores.}
    \label{fig:loss_matrix}
\end{figure*}

\clearpage

\section{Details of the Studied Languages}\label{appx:languages}
\begin{table*}[!h]
\small
    \centering
    \begin{tabular}{lllll}
    \toprule
    \textbf{Language Code} & \textbf{Name} & \textbf{Script} & \textbf{Family} & \textbf{NLLB resource-level} \\ 
        \midrule
        eus\_Latn & Basque & Latin & – & High \\ 
        hun\_Latn & Hungarian & Latin & – & High \\ 
        als\_Latn & Tosk Albanian & Latin & Albanian & High \\ 
        hye\_Armn & Armenian & Armenian & Armenic & Low \\ 
        bel\_Cyrl & Belarusian & Cyrillic & Balto-Slavic & Low \\ 
        srp\_Cyrl & Serbian & Cyrillic & Balto-Slavic & Low \\ 
        ukr\_Cyrl & Ukrainian & Cyrillic & Balto-Slavic & High \\ 
        mkd\_Cyrl & Macedonian & Cyrillic & Balto-Slavic & High \\ 
        bul\_Cyrl & Bulgarian & Cyrillic & Balto-Slavic & High \\ 
        rus\_Cyrl & Russian & Cyrillic & Balto-Slavic & High \\ 
        lvs\_Latn & Standard Latvian & Latin & Balto-Slavic & High \\ 
        bos\_Latn & Bosnian & Latin & Balto-Slavic & High \\ 
        lit\_Latn & Lithuanian & Latin & Balto-Slavic & High \\ 
        slk\_Latn & Slovak & Latin & Balto-Slavic & High \\ 
        slv\_Latn & Slovenian & Latin & Balto-Slavic & High \\ 
        hrv\_Latn & Croatian & Latin & Balto-Slavic & High \\ 
        ces\_Latn & Czech & Latin & Balto-Slavic & High \\ 
        pol\_Latn & Polish & Latin & Balto-Slavic & High \\ 
        xho\_Latn & Xhosa & Latin & Benue-Congo & High \\ 
        swh\_Latn & Swahili & Latin & Benue-Congo & High \\ 
        mya\_Mymr & Burmese & Myanmar & Burmo-Qiangic & ~ \\ 
        gla\_Latn & Scottish Gaelic & Latin & Celtic & Low \\ 
        cym\_Latn & Welsh & Latin & Celtic & Low \\ 
        gle\_Latn & Irish & Latin & Celtic & Low \\ 
        hau\_Latn & Hausa & Latin & Chadic & Low \\ 
        uig\_Arab & Uyghur & Arabic & Common Turkic & Low \\ 
        kir\_Cyrl & Kyrgyz & Cyrillic & Common Turkic & Low \\ 
        kaz\_Cyrl & Kazakh & Cyrillic & Common Turkic & High \\ 
        azj\_Latn & North Azerbaijani & Latin & Common Turkic & Low \\ 
        uzn\_Latn & Northern Uzbek & Latin & Common Turkic & High \\ 
        tur\_Latn & Turkish & Latin & Common Turkic & High \\ 
        som\_Latn & Somali & Latin & Cushitic & Low \\ 
        gaz\_Latn & West Central Oromo & Latin & Cushitic & Low \\ 
        epo\_Latn & Esperanto & Latin & Esperantic & Low \\ 
        est\_Latn & Estonian & Latin & Finnic & High \\ 
        fin\_Latn & Finnish & Latin & Finnic & High \\ 
        kat\_Geor & Georgian & Georgian & Georgian-Zan & Low \\ 
        ydd\_Hebr & Eastern Yiddish & Hebrew & Germanic & Low \\ 
        nno\_Latn & Norwegian Nynorsk & Latin & Germanic & Low \\ 
        isl\_Latn & Icelandic & Latin & Germanic & High \\ 
        afr\_Latn & Afrikaans & Latin & Germanic & High \\ 
        swe\_Latn & Swedish & Latin & Germanic & High \\ 
        dan\_Latn & Danish & Latin & Germanic & High \\ 
        deu\_Latn & German & Latin & Germanic & High \\ 
        nld\_Latn & Dutch & Latin & Germanic & High \\ 
        eng\_Latn & English & Latin & Germanic & High \\ 

        \bottomrule
    \end{tabular}
    \caption{Studied languages sorted by family (part 1).}
\end{table*}

\begin{table*}[!h]
    \centering
    \small
    \begin{tabular}{lllll}
    \toprule
        \textbf{Language Code} & \textbf{Name} & \textbf{Script} & \textbf{Family} & \textbf{NLLB resource-level} \\ 
        \midrule
        ell\_Grek & Greek & Greek & Graeco-Phrygian & High \\ 
        snd\_Arab & Sindhi & Arabic & Indo-Aryan & Low \\ 
        urd\_Arab & Urdu & Arabic & Indo-Aryan & Low \\ 
        asm\_Beng & Assamese & Bengali & Indo-Aryan & Low \\ 
        ben\_Beng & Bengali & Bengali & Indo-Aryan & High \\ 
        san\_Deva & Sanskrit & Devanagari & Indo-Aryan & Low \\ 
        npi\_Deva & Nepali & Devanagari & Indo-Aryan & Low \\ 
        mar\_Deva & Marathi & Devanagari & Indo-Aryan & Low \\ 
        hin\_Deva & Hindi & Devanagari & Indo-Aryan & High \\ 
        guj\_Gujr & Gujarati & Gujarati & Indo-Aryan & Low \\ 
        pan\_Guru & Eastern Panjabi & Gurmukhi & Indo-Aryan & Low \\ 
        ory\_Orya & Odia & Oriya & Indo-Aryan & Low \\ 
        sin\_Sinh & Sinhala & Sinhala & Indo-Aryan & Low \\ 
        pbt\_Arab & Southern Pashto & Arabic & Iranian & Low \\ 
        prs\_Arab & Dari & Arabic & Iranian & Low \\ 
        kmr\_Latn & Northern Kurdish & Latin & Iranian & Low \\ 
        glg\_Latn & Galician & Latin & Italic & Low \\ 
        cat\_Latn & Catalan & Latin & Italic & High \\ 
        spa\_Latn & Spanish & Latin & Italic & High \\ 
        por\_Latn & Portuguese & Latin & Italic & High \\ 
        ita\_Latn & Italian & Latin & Italic & High \\ 
        fra\_Latn & French & Latin & Italic & High \\ 
        ron\_Latn & Romanian & Latin & Italic & High \\ 
        jpn\_Jpan & Japanese & Japanese & Japanesic & High \\ 
        lao\_Laoo & Lao & Lao & Kam-Tai & Low \\ 
        tha\_Thai & Thai & Thai & Kam-Tai & High \\ 
        khm\_Khmr & Khmer & Khmer & Khmeric & Low \\ 
        kor\_Hang & Korean & Hangul & Korean & High \\ 
        sun\_Latn & Sundanese & Latin & Malayo-Polynesian & Low \\ 
        jav\_Latn & Javanese & Latin & Malayo-Polynesian & Low \\ 
        plt\_Latn & Plateau Malagasy & Latin & Malayo-Polynesian & Low \\ 
        zsm\_Latn & Standard Malay & Latin & Malayo-Polynesian & High \\ 
        ind\_Latn & Indonesian & Latin & Malayo-Polynesian & High \\ 
        khk\_Cyrl & Halh Mongolian & Cyrillic & Mongolic & Low \\ 
        arb\_Arab & Modern Standard Arabic & Arabic & Semitic & High \\ 
        amh\_Ethi & Amharic & Geez & Semitic & Low \\ 
        heb\_Hebr & Hebrew & Hebrew & Semitic & High \\ 
        zho\_Hans & Chinese & Han (Simplified) & Sinitic & ~ \\ 
        zho\_Hant & Chinese & Han (Traditional) & Sinitic & ~ \\ 
        kan\_Knda & Kannada & Kannada & South Dravidian & ~ \\ 
        mal\_Mlym & Malayalam & Malayalam & South Dravidian & ~ \\ 
        tam\_Taml & Tamil & Tamil & South Dravidian & ~ \\ 
        tel\_Telu & Telugu & Telugu & South Dravidian & ~ \\ 
        vie\_Latn & Vietnamese & Latin & Vietic & High \\ 
        \bottomrule
    \end{tabular}
    \caption{Studied languages sorted by family (part 2).}
\end{table*}

\clearpage
\section{Embedding Similarity Additional Results}
\label{appx:embedding_similarity}
\label{appx:matrices}
\appendix
\begin{figure*}[!h]
    \centering
    \includegraphics[width=\linewidth]{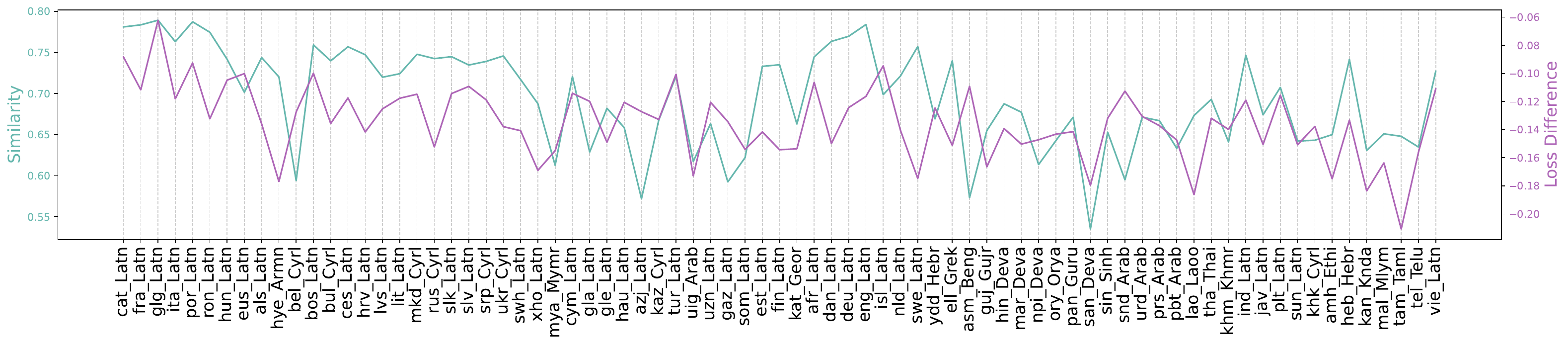}
    \caption{Comparison between the row corresponding to Spanish of the interference matrix and SONAR similarity.}
\end{figure*}
\begin{figure*}[!h]
    \centering
    \includegraphics[width=\linewidth]{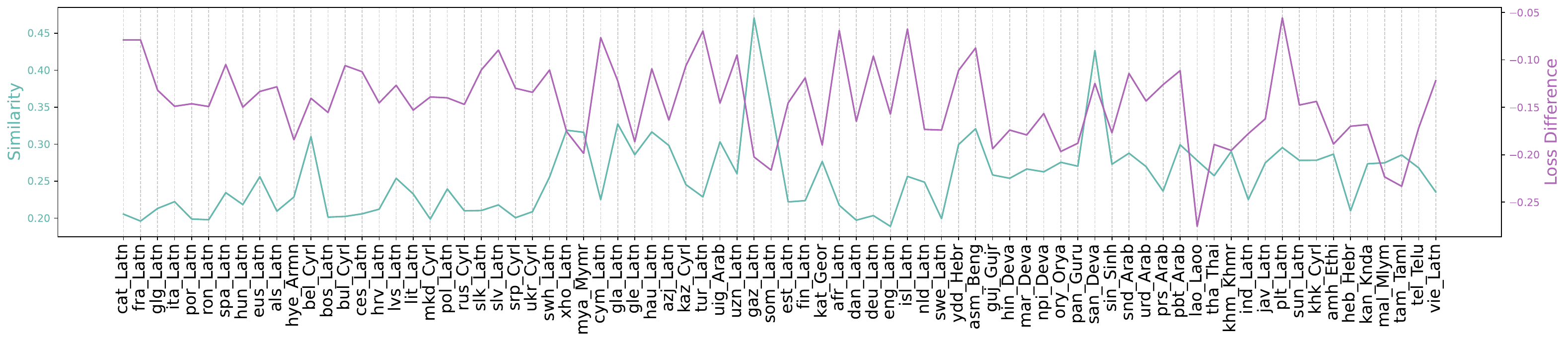}
    \caption{Comparison between the row corresponding to Greek of the interference matrix and MEXMA similarity.}
\end{figure*}
\begin{figure*}[!h]
    \centering
    \includegraphics[width=\linewidth]{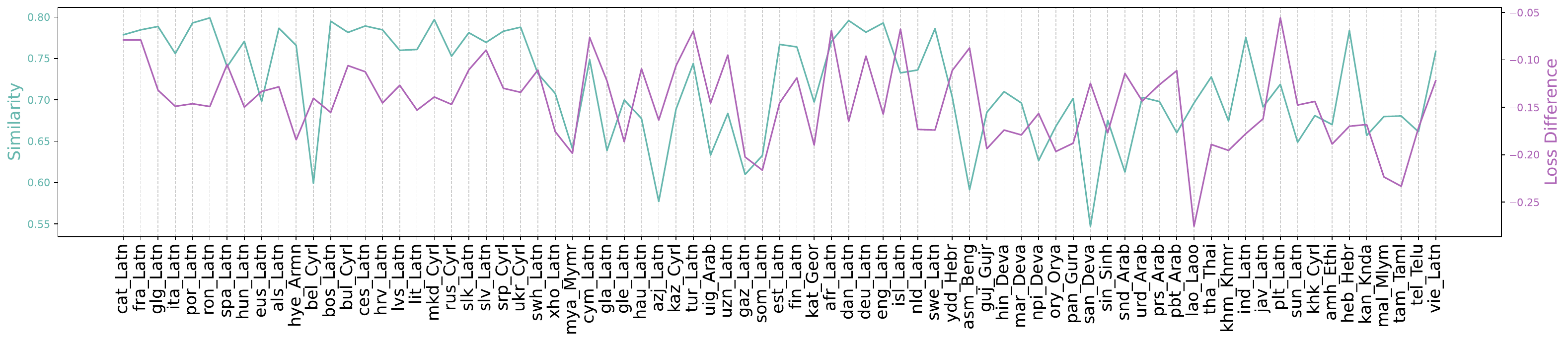}
    \caption{Comparison between the row corresponding to Greek of the interference matrix and SONAR similarity.}
\end{figure*}
\begin{figure*}[!h]
    \centering
    \includegraphics[width=\linewidth]{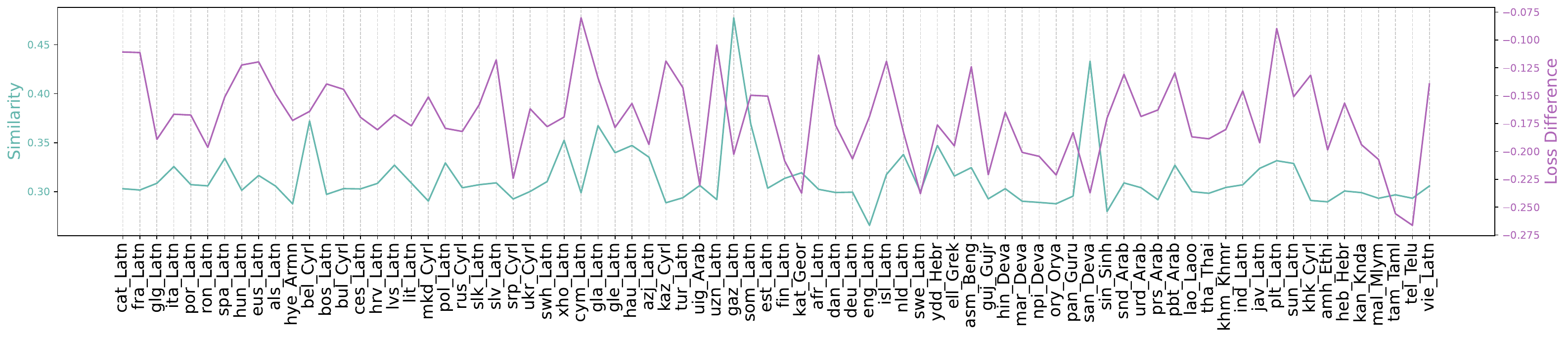}
    \caption{Comparison between the row corresponding to Burmese of the interference matrix and MEXMA similarity.}
\end{figure*}
\begin{figure*}[!h]
    \centering
    \includegraphics[width=\linewidth]{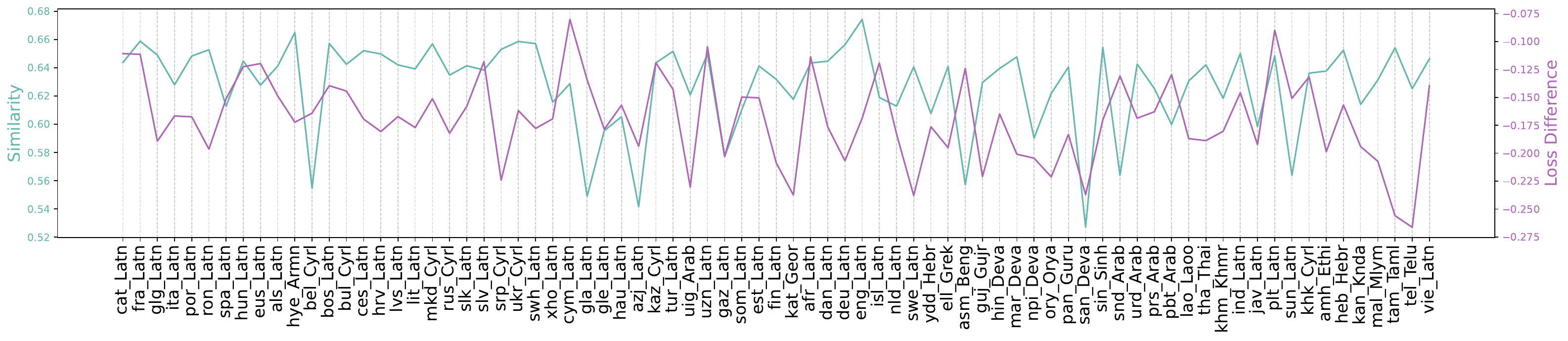}
    \caption{Comparison between the row corresponding to Burmese of the interference matrix and SONAR similarity.}
\end{figure*}
\begin{figure*}[!h]
    \centering
    \includegraphics[width=\linewidth]{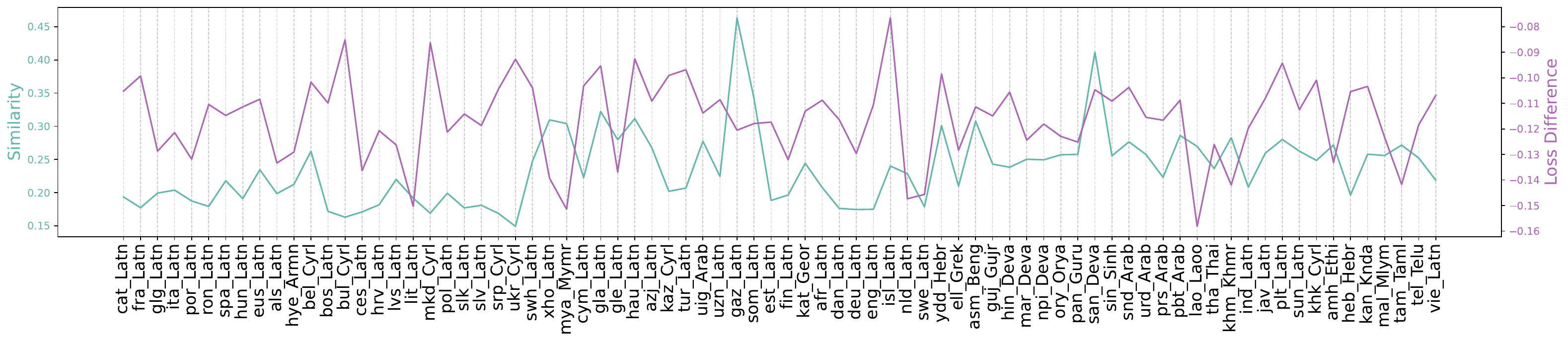}
    \caption{Comparison between the row corresponding to Russian of the interference matrix and MEXMA similarity.}
\end{figure*}
\begin{figure*}[!h]
    \centering
    \includegraphics[width=\linewidth]{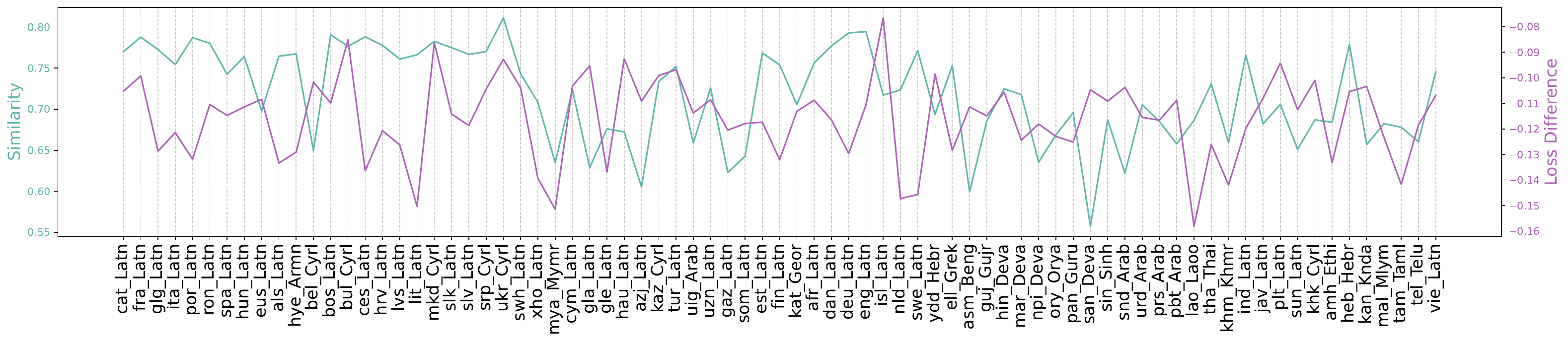}
    \caption{Comparison between the row corresponding to Russian of the interference matrix and SONAR similarity.}
\end{figure*}
\begin{figure*}[!h]
    \centering
    \includegraphics[width=\linewidth]{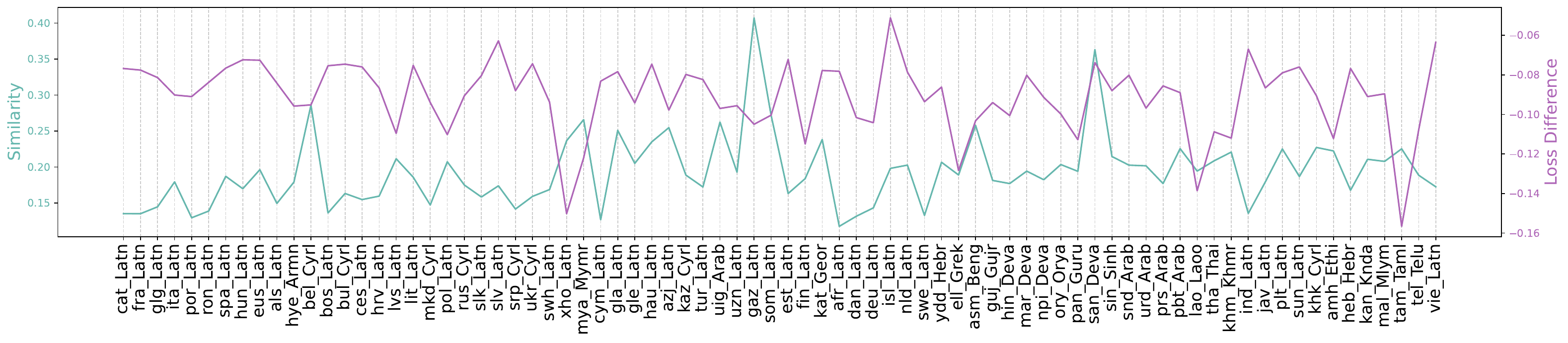}
    \caption{Comparison between the row corresponding to English of the interference matrix and MEXMA similarity.}
\end{figure*}
\begin{figure*}[!h]
    \centering
    \includegraphics[width=\linewidth]{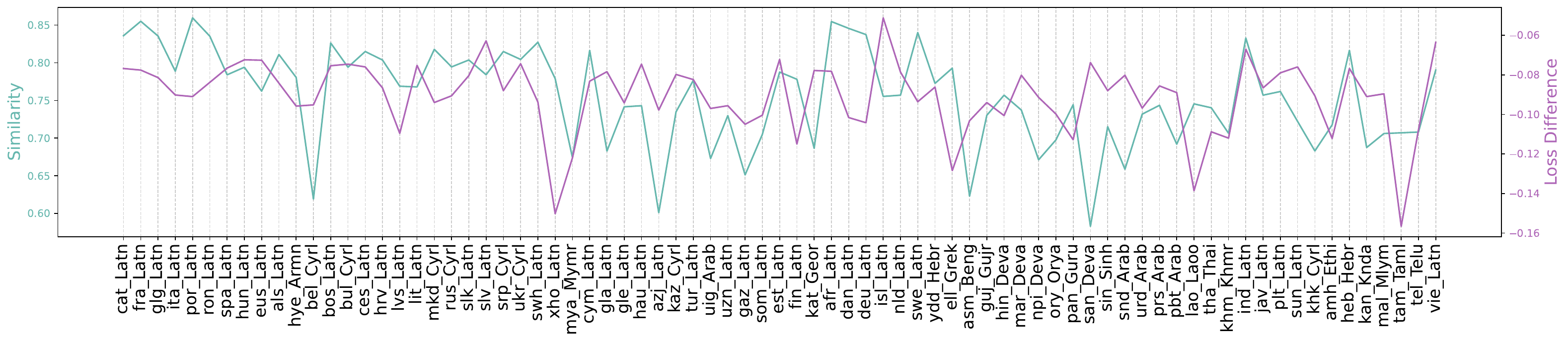}
    \caption{Comparison between the row corresponding to English of the interference matrix and SONAR similarity.}
\end{figure*}

\end{document}